\documentclass{article}

\usepackage[table,svgnames,dvipsnames]{xcolor}
\usepackage{amsmath,amssymb,amsfonts}
\usepackage{algorithmic}
\usepackage{graphicx}
\usepackage{textcomp}
\usepackage{booktabs}
\usepackage{pgfplots}
\usepgfplotslibrary{groupplots}
\pgfplotsset{compat=1.18}
\usepackage{multirow}
\usepackage[caption=false,font=footnotesize]{subfig}
\usepackage{todonotes}
\usepackage[export]{adjustbox}
\usepackage{tabularx}
\usepackage{enumitem}
\setlist[itemize]{noitemsep}
\usepackage[font=small,labelfont=bf]{caption}
\usepackage{acra}
\usepackage{tikz}
\usetikzlibrary{positioning,calc}
\usepackage[flushleft]{threeparttable}
\usepackage[normalem]{ulem}
\usepackage{xcolor}
\usepackage{fancyhdr}

\author{
  M. Oltan Sevinc$^{1}$, Liao Wu$^{2}$, Francisco Cruz$^{1, 3}$\\
  $^{1}$School of Computer Science and Engineering, University of New South Wales, Australia\\
  $^{2}$School of Mechanical and Manufacturing Engineering, University of New South Wales, Australia\\
  $^{3}$Escuela de Ingeniería, Universidad Central de Chile, Santiago, Chile\\
  \texttt{\{m.sevinc, liao.wu, f.cruz\}@unsw.edu.au}
}
\title
{
    Towards Closing the Domain Gap with Event Cameras\\
} 

\fancypagestyle{fancytitle}{
  \fancyhf{} 
  \fancyhead[L]{\small Please cite as: M. Oltan Sevinc, Liao Wu, and Francisco Cruz, ``Towards Closing the Domain Gap with Event Cameras'', \textit{Proceedings of the Australasian Conference on Robotics and Automation (ACRA), 2025}.}
  
}

\begin{document}
\maketitle
\thispagestyle{fancytitle}
\begin{abstract}
  Although traditional cameras are the primary sensor for end-to-end driving, their performance suffers greatly when the conditions of the data they were trained on does not match the deployment environment, a problem known as the domain gap. In this work, we consider the day-night lighting difference domain gap. Instead of traditional cameras we propose event cameras as a potential alternative which can maintain performance across lighting condition domain gaps without requiring additional adjustments. Our results show that event cameras maintain more consistent performance across lighting conditions, exhibiting domain-shift penalties that are generally comparable to or smaller than grayscale frames and provide superior baseline performance in cross-domain scenarios.

\end{abstract}
\section{Introduction} \label{introduction-section}
End-to-end driving is a holistic approach to autonomous driving, where sensor data is directly used to generate control outputs. It stands in contrast to module-by-module optimized approaches, where aspects of a vehicle control task are broken down into human-understandable chunks and tackled one at a time, using methods such as neural networks~\cite{chibRecentAdvancementsEndtoEnd2024}.

End-to-end approaches may be preferable to module-by-module optimized approaches as they~\cite{chenEndtoEndAutonomousDriving2024}:
\begin{itemize}[topsep=0pt]
    \item Allow the loss function to propagate end-to-end, and therefore ensure there are no module losses that do not line up with the end goal.
    \item Scale well with data-driven approaches, as the entire system can improve with more data.
    \item Perform data fusion intuitively.
\end{itemize}

Frame-based cameras are well established as the go-to sensor for end-to-end driving~\cite{bojarskiEndEndLearning2016,chenLearningCheating2020,zhangEndtoEndUrbanDriving2021,wuTrajectoryguidedControlPrediction2022}. However, neural networks trained on frame-based camera data are often troubled by the domain gap problem, a degradation of performance when aspects of the training set, such as lighting conditions or environmental characteristics do not match those of the testing environment~\cite{schuteraNighttoDayOnlineImagetoImage2021,sunSeeClearerNight2019}.

Unlike standard cameras, which capture absolute intensity frames, event cameras operate on pixel-level relative brightness changes~\cite{rebecqRealtimeVisualInertialOdometry2017}. We hypothesize that this quality reduces any domain gap based on lighting difference, and leads to less performance degradation compared to a frame-based camera when a neural network trained on event camera data encounters novel lighting conditions. This illumination invariance is particularly valuable for autonomous robotic systems that must operate reliably across varying environmental conditions without human intervention.

To evaluate this hypothesis, we train end-to-end driving models on datasets biased toward daytime and nighttime driving for two sensor modalities: a conventional grayscale camera and a framed event camera. We evaluate each model on both daytime and nighttime test sets and quantify the performance degradation incurred when models are applied to out-of-domain illumination conditions.

Proceeding from this hypothesis, our contributions are as follows:
\begin{itemize}[topsep=0pt]
    \item We inspect the domain shift penalty encountered by grayscale and framed event camera recordings. We show that a neural network trained on a day-biased event camera dataset strongly outperforms its grayscale counterpart in night conditions.
    \item We compare grayscale and framed event camera recordings from day and night to show event-camera data profiles remain consistent across lighting conditions.
\end{itemize}
\section{Related Work} \label{relatedwork-section}
Behavior cloning~\cite{bainFrameworkBehaviouralCloning2000}, a branch of imitation learning, has gained prominence as a method for tackling the end-to-end driving problem. Behavior cloning involves the application of supervised learning over a dataset collected by a human expert solving the problem. It is an enticing approach because it simplifies the problem to the well-studied supervised learning one, with a clear loss signal. Behavior cloning has a rich history dating back to ALVINN~\cite{pomerleauALVINNAutonomousLand1988}, which used a rudimentary neural network to steer a vehicle. The modern era of neural network-powered end-to-end driving took off after Bojarski et al.~\cite{bojarskiEndEndLearning2016} demonstrated the potential modern deep-learning powered neural network approaches present in this domain, using a convolutional neural network with standard camera input.

Event cameras are another candidate sensor for end-to-end driving. An initial dataset and attempt at event camera-based end-to-end behavior cloning for driving came in DDD17~\cite{binasDDD17EndToEndDAVIS2017}. DDD17 used dynamic vision sensors (DVSs, a subset of event cameras) in conjunction with an active pixel sensor (APS) generating grayscale frames, for steering angle regression. They aggregated their DVS events into frames, inspired by the work of~\cite{moeysSteeringPredatorRobot2016}. An aggregation method was necessary, as event cameras are asynchronous in nature and thus cannot be readily processed by established methods in neural network literature, without first being converted to a format that parallelizes them into a compatible representation. This was taken a step further by Maqueda et al.~\cite{maquedaEventBasedVisionMeets2018} who introduced a sophisticated event framing method based on 2D histograms of ON and OFF events. Maqueda et al. then successfully used this method on the DDD17 dataset to establish a baseline for event frame-based end-to-end driving. A large extension to the DDD17 dataset was presented in DDD20~\cite{huDDD20EndtoEndEvent2020}.

A biased dataset can negatively affect the performance of any behavior cloning model significantly~\cite{codevillaExploringLimitationsBehavior2019}. For frame-based cameras, one of the sources of this bias can be the lighting conditions~\cite{schuteraNighttoDayOnlineImagetoImage2021,chenEndtoEndAutonomousDriving2024}, with many datasets focused on more idealized daytime lighting conditions~\cite{daiDarkModelAdaptation2018}. One approach to address this problem is lighting-invariant transforms of RGB images~\cite{maddernIlluminationInvariantImaging2014}. Another approach is the real-time conversion of night images into the day domain better known to the network~\cite{schuteraNighttoDayOnlineImagetoImage2021}.

Event cameras have shown their promise across different lighting conditions, with their ability to detect small relative lighting changes being applied to enhance dark images~\cite{liuSeeingMotionNighttime2024} as well as for performing object detection across varying lighting conditions~\cite{caoChasingDayNight2024}. We find the next logical step is to utilize this potential to perform across varying illumination conditions to tackle the biased dataset problem, a connection that to our knowledge has not previously been made in literature.

\section{Proposed Approach} \label{approach-section}
Working towards our goal of evaluating the lighting invariance of event cameras compared to grayscale cameras in an end-to-end driving task, we:
\begin{enumerate}
  \item select a dataset that captures driving data across different lighting conditions using both types of cameras,
  \item propose a preprocessing and data augmentation pipeline to improve the viability of the dataset for the comparison,
  \item train separate end-to-end driving models for each sensor type using training data biased towards either daytime or nighttime samples, and
  \item evaluate each trained model on test sets composed exclusively of daytime or exclusively of nighttime samples.
\end{enumerate}

\subsection{Dataset}
For our experiments we use the DDD20 dataset, introduced in Section~\ref{relatedwork-section}. This dataset presents driving footage captured by the DAVIS~\cite{brandli2401801302014} sensor array, the vehicle's steering angle, and other data captured across different lighting conditions. The DAVIS sensor array is made of an APS along with a DVS. The APS has an average refresh rate of 50~ms. We choose the vehicle steering angle as the output variable we want our model to learn.

Using the utilities provided with~\cite{huDDD20EndtoEndEvent2020}, we aggregate the event camera data in the dataset by combining 2D histograms of ON and OFF events over a period of 50~ms, creating event frames. The 50~ms period temporally matches the APS, and it was shown to be the most performant integration period for the task in~\cite{maquedaEventBasedVisionMeets2018}.
We then select full recordings from each of the dataset's daytime and nighttime recordings, yielding approximately 130,000 matching APS and DVS frames for each lighting condition. The full recordings are split into their first 85\% and their last 15\%, assigning the initial 85\% of each recording as training data, and keeping their remaining 15\% as testing data. We create two datasets from our selected training frames, \textit{day biased} and \textit{night biased}. Each biased dataset contains all samples from its target subset and at most 25\% worth of samples from the opposite subset (i.e., a maximum approximate imbalance of 80:20). An illustration of our data split can be seen in Figure~\ref{fig:data-split}.

{
\begin{figure*}[t]
  \centering
  \includegraphics[height=225px]{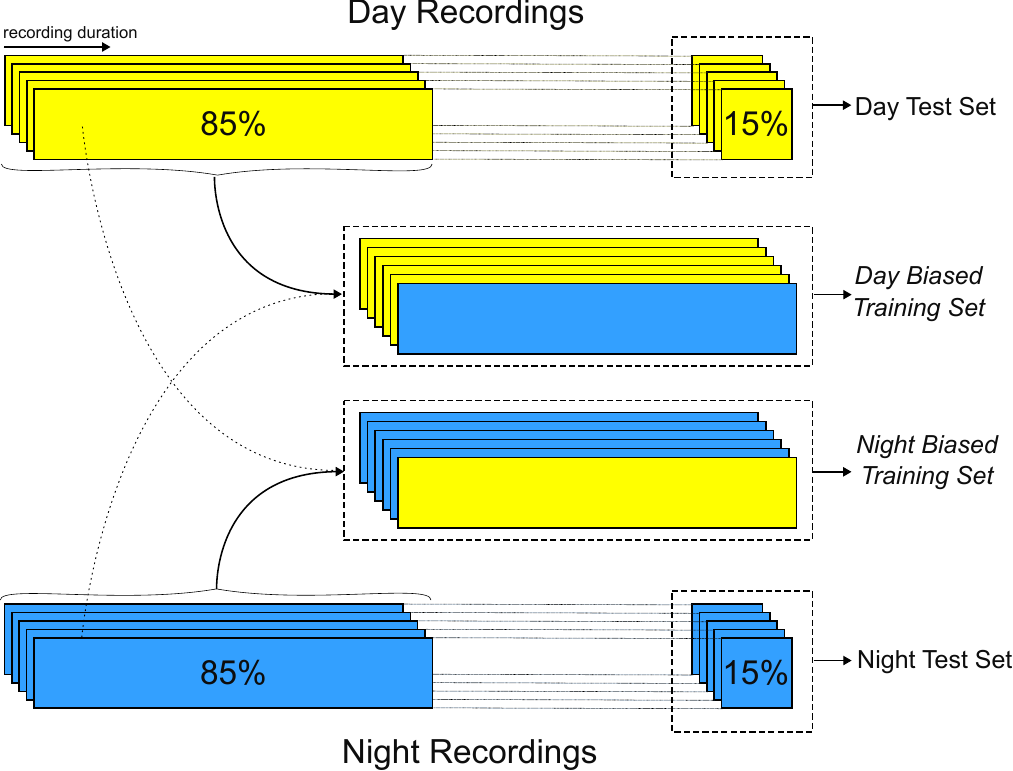}
  \caption{The data split methodology we use to create our day biased and night biased training sets. Our process creates training sets that intentionally suffer from a domain gap, while our test sets are purely from one lighting condition, to test the models' ability to generalize across the domain gap.}
  \label{fig:data-split}
\end{figure*}
}

We proceed to preprocess our data, including pruning as described in Section~\ref{ssection-preprocessing}. Following pruning, the train/test split of our dataset is 70.1\%/29.9\%.

\subsection{Preprocessing}\label{ssection-preprocessing}
\begin{figure*}[t]
  \centering
  \includegraphics[max width=\textwidth, max height=275px]{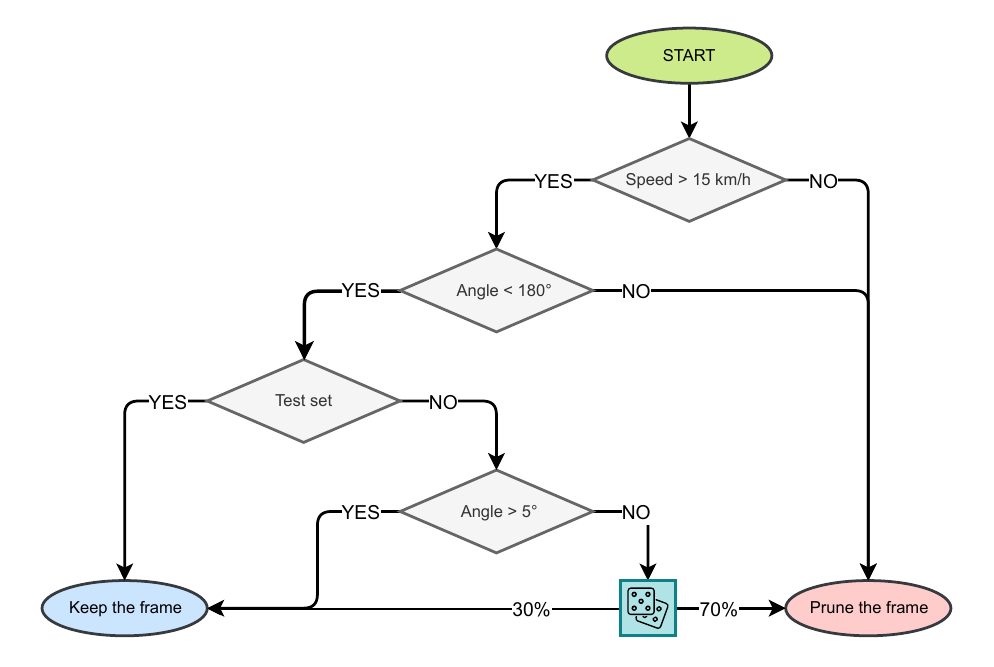}
  \caption{Flowchart of the pruning process, excludes manual trimming of the beginning and end of the recordings.}
  \label{fig:flowchart}
\end{figure*}

As a part of our preprocessing of the dataset, we prune certain parts of it to improve its usefulness and to remove outliers. A flowchart of this pruning process can be seen in Figure~\ref{fig:flowchart}. We adapt the majority of our preprocessing pipeline from~\cite{huDDD20EndtoEndEvent2020}, and note when we deviate. The descriptions and justifications of our preprocessing procedure are as such:

\textit{Pruning Low-Speed Frames}: In driving datasets, low-speed scenarios often correspond to situations like junctions and traffic lights, and the driver's behavior at these points cannot be predicted even by an observing person. Thus, we prune the frames with a speed below 15 km/h.

\textit{Pruning Low Steering Angle Frames}: In everyday scenarios, roads are often approximately straight, with low steering angles to match. This can cause a strong bias in the data towards low steering angles, and encourages models to guess the mean of the data. Hence, as the next preprocessing step we randomly prune 70\% of the frames with an absolute steering angle less than 5 degrees.

\textit{Pruning Extreme Steering Angle Frames}: Similar to our rule associated with low speeds, virtually any scenario where the steering wheel would be turned by over 180° in either direction, excluding hairpin turns, represents a situation where the model would need to develop prescience in order to predict. Thus, we prune any frames that have an absolute steering angle over 180°. This represents a deviation from~\cite{huDDD20EndtoEndEvent2020}, which trim the steering angles at three times standard deviation instead.

\textit{Pruning Beginning/Ending of Recordings}: Although it is expected that they would fall within the bounds of the pruning criteria described before, we also manually prune the beginning and ending of recordings where the driver is pulling out of/into roads.

\textit{Rescaling Steering Angles}: Empowered by our choice to map our minimum and maximum steering angles to the range [-180, 180], we scale our target steering angle range to $[-1, 1]$ to encourage better training dynamics in another deviation from the prescribed pipeline in~\cite{huDDD20EndtoEndEvent2020}.

\textit{Normalizing APS and DVS Frames}: The APS and framed DVS data have a natural range of $[0, 255]$. Much like~\cite{maquedaEventBasedVisionMeets2018,huDDD20EndtoEndEvent2020}, we normalize both the APS and DVS frames to the interval $[0,1]$ by dividing all pixel values by 255 to encourage better training dynamics.

\textit{Resizing}: We resize our DVS and APS frames to the size $224 \times 224$ pixels from their original size of $346 \times 260$ pixels (using bilinear interpolation), to enable direct compatibility with commonly used neural network architectures, unlike the $172 \times 128$ resolution in~\cite{huDDD20EndtoEndEvent2020}.

\subsection{Data Augmentation} \label{subsection:augmentation}
We augment the training data with random transformations to improve the network's ability to generalize. These empirically determined random transformations are:
\begin{itemize}[topsep=0pt]
  \item $30\%$ chance of a random resized crop, with a width-to-height ratio between $[0.8, 1.2]$, and a scale between $[0.8, 1.0]$.
  \item $40\%$ chance of a random rotation, of up to $3$ degrees.
  \item $20\%$ chance of a color jitter with a brightness of $0.2$ and a contrast of $0.2$.
  \item $30\%$ chance of adding random Gaussian noise, with $\mu=0$ and $\sigma=0.01$.
  \item $20\%$ chance of a Gaussian blur, with a kernel size of $3$.
\end{itemize}

\subsection{Training Details}
\begin{figure*}[htbp]
  \centering
  \includegraphics[height=275px]{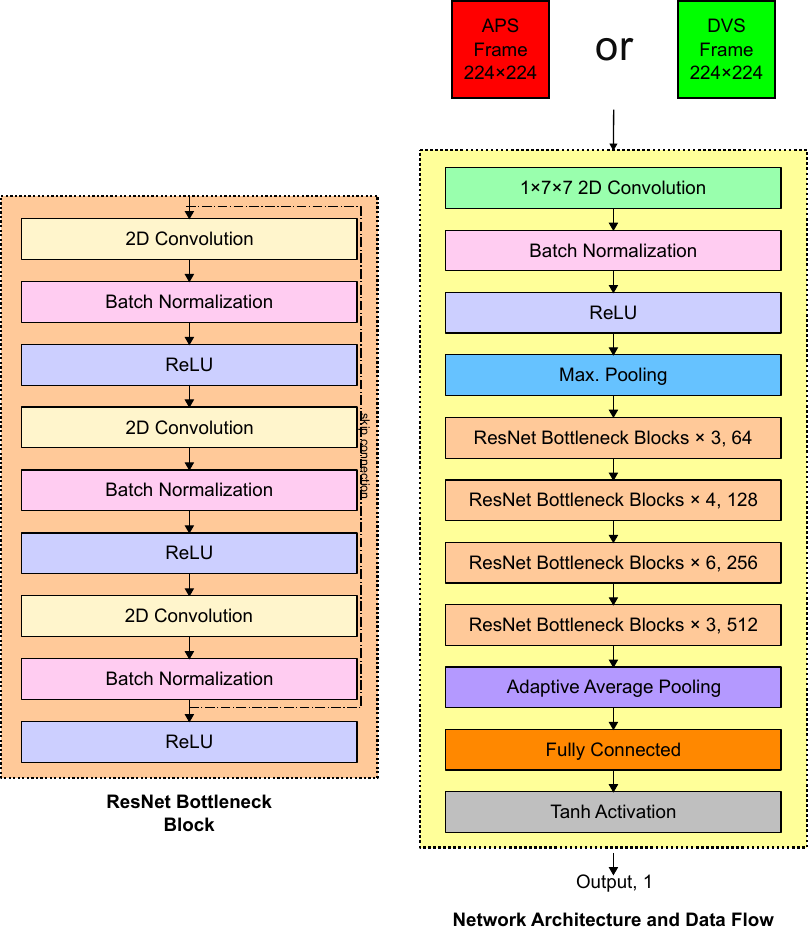}
  \caption{Our modified Resnet-50 architecture, with an ImageNet initialization. The input layer is modified to accept 1-channel images, and a Tanh activation function is added to the head to match the $[-1, 1]$ range of the scaled steering angle output.}
  \label{fig:network-architecture}
\end{figure*}

To get our results, we utilize a Resnet-50~\cite{heDeepResidualLearning2015}, with an ImageNet~\cite{dengImageNetLargescaleHierarchical2009} initialization as done in~\cite{maquedaEventBasedVisionMeets2018}, modified to take in 1-channel images and with a Tanh activation function added to its head to exploit the $[-1, 1]$ range the output data has been scaled to in the preprocessing step outlined in Section~\ref{ssection-preprocessing}. An image of our network can be seen in Figure~\ref{fig:network-architecture}. We use the AdamW~\cite{loshchilovDecoupledWeightDecay2019} optimizer with an initial learning rate of $3e-4$ and a weight decay of $1e-3$. For each model, we pick the epoch with the highest test set performance.

\subsection{Metrics} \label{metrics}
We match the metrics used in the original presentation of the dataset~\cite{huDDD20EndtoEndEvent2020}, root mean squared error (RMSE) and explained variance (shortened to EVA here to match~\cite{huDDD20EndtoEndEvent2020,chenGFENetGroupwiseFeatureenhanced2024} but often $R^2$ in nomenclature), to ensure the comparability of our results.

We use the RMSE, Equation~(\ref{eq:RMSE}), of the steering angles as our loss signal when training our models. In Equation~(\ref{eq:RMSE}), $y_i$ denotes the ground-truth steering angle for sample $i$, $\hat{y}_i$ the model prediction for that sample, and $n$ the number of samples. Lower values indicate better RMSE, with the lowest possible value being 0.

\begin{equation}
  \text{RMSE} = \sqrt{\frac{1}{n} \sum_{i=1}^{n} (y_i - \hat{y}_i)^2}\label{eq:RMSE}
\end{equation}

We present EVA, Equation~(\ref{eq:EVA}), as an auxiliary measure that demonstrates that our model predictions are not merely the means of the data they are trained on, and instead depart from their mean as the ground truth does. In Equation~(\ref{eq:EVA}), $\mathrm{Var}(\hat{y}-y)$ denotes the sample variance of the residuals $(\hat{y}_i - y_i)$ across the dataset, and $\mathrm{Var}(y)$ denotes the sample variance of the ground-truth steering angles $y_i$. EVA ranges up to 1 (perfect prediction); a value of 0 indicates the model always predicts the mean of $y$, and negative values indicate worse performance than always predicting the mean.

\begin{equation}
  \text{EVA} = 1 - \frac{\text{Var}(\hat{y} - y)}{\text{Var}(y)}\label{eq:EVA}
\end{equation}
\section{Experimental Results and Discussion}\label{experimentsresults-section}
To justify our domain-shift experiments, we characterize the APS and DVS data across the daytime and nighttime portions of our dataset. The data presented in Table~\ref{tab:aps_dvs_day_night_stats} numerically demonstrates that while the APS data profile changes drastically across lighting conditions, the DVS data maintains a more consistent profile. We further quantify this consistency using relative changes in statistics to assess data variability and Cohen's d effect size to measure the standardized difference between day and night distributions. Cohen's d values of 0.2, 0.5, and 0.8 represent small, medium, and large effect sizes, respectively.

\begin{figure*}[htbp]
  \centering
  \def\imgheight{0.65\textheight}

  \subfloat[DVS Day]{%
    \includegraphics[height=\imgheight]{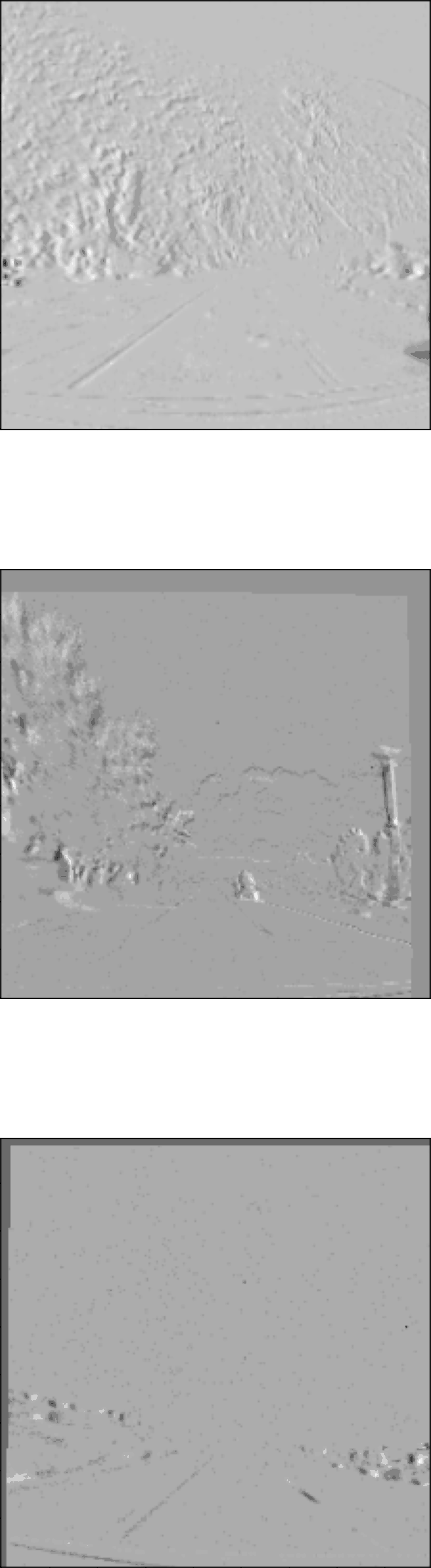}%
  }
  \hfill
  \subfloat[DVS Night]{%
    \includegraphics[height=\imgheight]{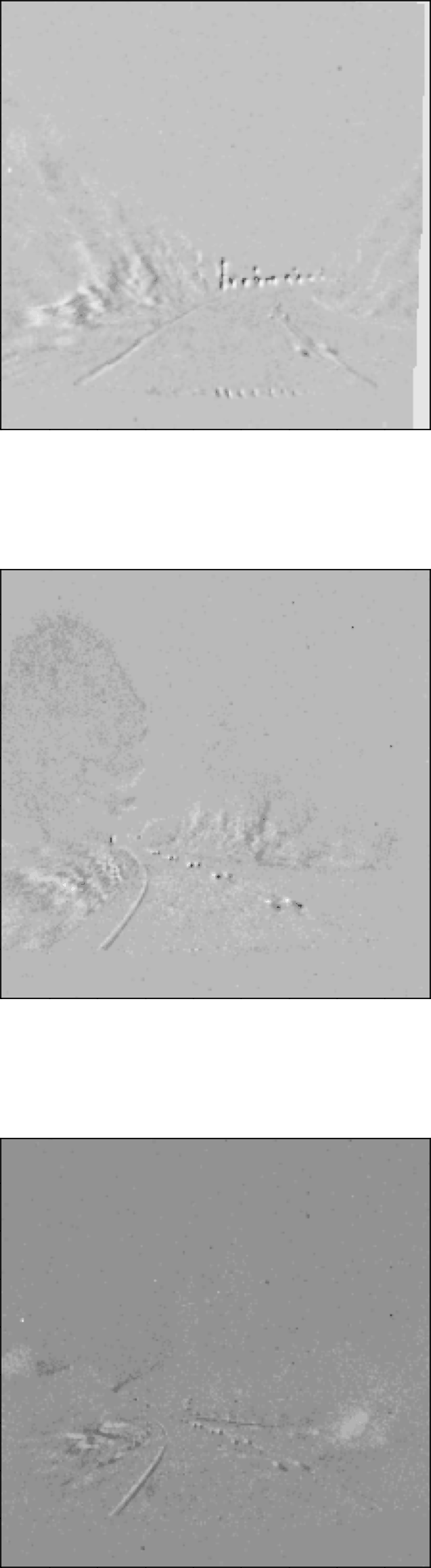}%
  }
  \hfill
  \rule{0.3pt}{\imgheight}%
  \hfill
  \subfloat[APS Day]{%
    \includegraphics[height=\imgheight]{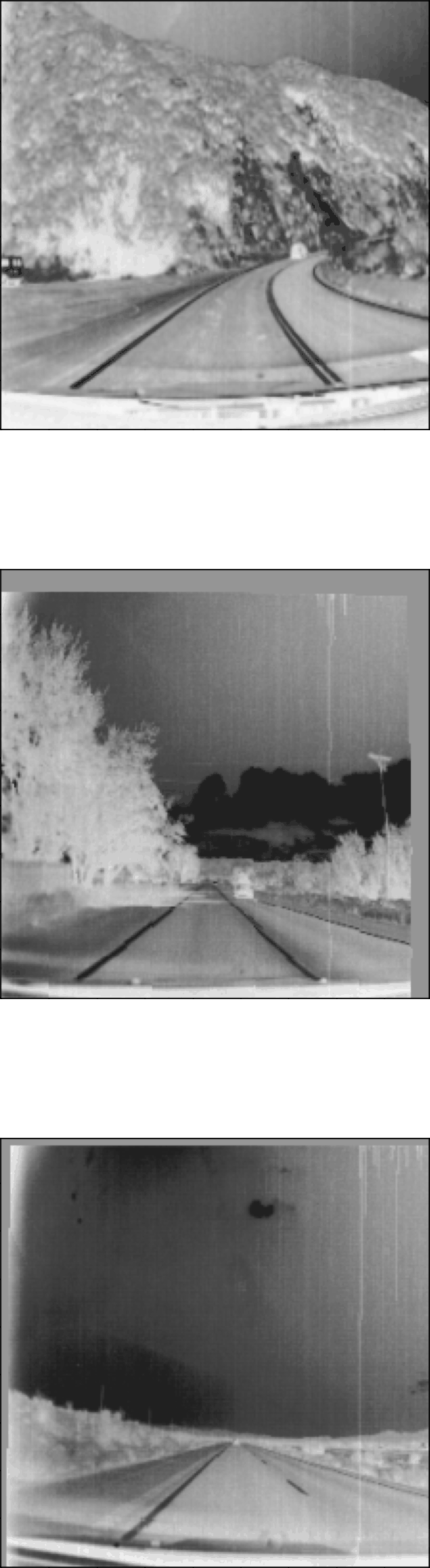}%
  }
  \hfill
  \subfloat[APS Night]{%
    \includegraphics[height=\imgheight]{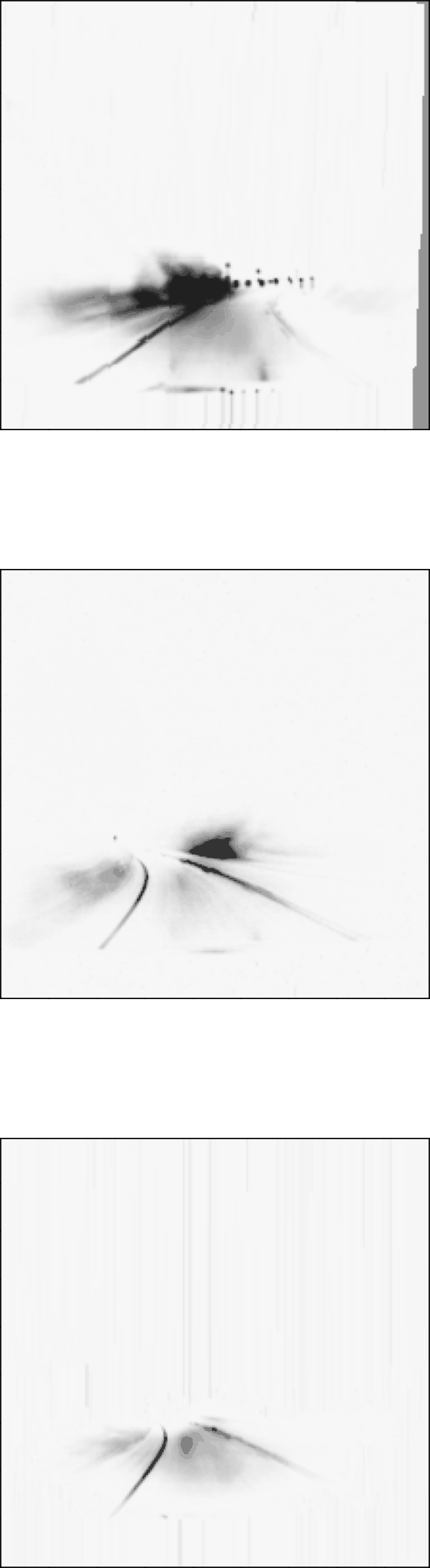}%
  }
  \caption{Comparison of DVS and APS sensor data under different lighting conditions. (a) and (c), as well as (b) and (d) were captured simultaneously. The day and night images were chosen to match similar sections of road, with the first row showing a road near a hillside, the second row passing groups of trees on the left, and the third row featuring open highway. Inspecting (a) and (b), it is seen that the event-framed DVS data has little difference across the lighting conditions. This parallel is not repeated between (c) and (d), which diverge significantly. This is consistent with the day/night data profiles of the sensors as shown in Table \ref{tab:aps_dvs_day_night_stats}.}
  \label{fig:dvs_aps_comparison}
\end{figure*}

Cohen's d is calculated using Equation~\ref{eq:cohens_d}:
\begin{equation}
  d = \frac{\mu_{\text{day}} - \mu_{\text{night}}}{\sqrt{\frac{\sigma_{\text{day}}^2 + \sigma_{\text{night}}^2}{2}}}
  \label{eq:cohens_d}
\end{equation}
This simplified formula assumes equal sample sizes between day and night datasets, which is a reasonable approximation given our similar sample counts ($n_{\text{day}} \approx n_{\text{night}}$). We report the signed $d$ (positive means a larger day mean). Figure~\ref{fig:dvs_aps_comparison} visually justifies our experiments examining the lighting invariance of DVS's, showing minimal difference between daytime and nighttime DVS frames taken in similar locations, while the APS frames diverge significantly.

\begin{table}[htbp]
  \begin{threeparttable}
    \caption{APS and DVS statistics for day and night datasets. The DVS shows substantially smaller domain shift as measured by relative changes across lighting conditions and Cohen's d.}
    \label{tab:aps_dvs_day_night_stats}
    \centering
    \small
    \setlength{\tabcolsep}{4pt}
    \begin{tabular*}{\columnwidth}{@{\extracolsep{\fill}}lrrrr@{}}
      \toprule
      Sensor & \multicolumn{1}{r}{Day $\mu \pm \sigma$} & \multicolumn{1}{r}{Night $\mu \pm \sigma$} & \multicolumn{1}{r}{$\Delta\mu$ (\%)} & \multicolumn{1}{r}{Cohen's d} \\
      \midrule
      DVS    & $111.6 \pm 35.7$   & $103.8 \pm 24.4$     & $-7.0$           & 0.25      \\
      APS    & $159.6 \pm 83.8$   & $13.9 \pm 40.5$      & $-91.3$          & 2.21      \\
      \bottomrule
    \end{tabular*}
    \begin{tablenotes}
      \small
      \item Cohen's d calculations (Equation~\ref{eq:cohens_d}) assume day and night samples represent comparable road sections with lighting as the primary variable, and use a simplified formula assuming approximately equal sample sizes. These measures indicate general trends rather than precise quantifications of lighting effects in isolation.
    \end{tablenotes}
  \end{threeparttable}
\end{table}

We train separate models for each sensor (DVS and APS) using either the \textit{day biased} and \textit{night biased} training sets; we then evaluate each of these models on test sets composed exclusively of daytime or nighttime samples.
The results of this experiment are presented in Table~\ref{tab:results-raw}. They show that the DVS outperformed the APS across all trials, a result consistent with previous literature~\cite{maquedaEventBasedVisionMeets2018,chenGFENetGroupwiseFeatureenhanced2024}. Interestingly, the day-biased DVS outperforms the night-biased APS in night conditions for the task in both RMSE and EVA.

\begin{table}[htbp]
  \centering
  \small
  \setlength{\tabcolsep}{5pt}
  \caption{Steering regression performance by lighting condition, sensor, and training bias (lower RMSE, higher EVA are better).}
  \label{tab:results-raw}
  \begin{tabular*}{\columnwidth}{@{\extracolsep{\fill}}lllrr@{}}
    \toprule
    Lighting               & Sensor               & Train bias & RMSE    & EVA    \\
    \midrule
    \multirow{4}{*}{Day}   & \multirow{2}{*}{DVS} & Day        & 11.60   & 0.698  \\
    &                      & Night      & 17.30   & 0.327  \\
    \cmidrule{2-5}
    & \multirow{2}{*}{APS} & Day        & 16.49   & 0.388  \\
    &                      & Night      & 19.19   & 0.172  \\
    \midrule
    \multirow{4}{*}{Night} & \multirow{2}{*}{DVS} & Day        & 11.81   & 0.685  \\
    &                      & Night      & 8.10    & 0.852  \\
    \cmidrule{2-5}
    & \multirow{2}{*}{APS} & Day        & 18.07   & 0.263  \\
    &                      & Night      & 12.55   & 0.645  \\
    \bottomrule
  \end{tabular*}
\end{table}

While adding a small amount of data from the opposite lighting condition preserved acceptable cross-domain performance for the DVS, models trained exclusively on one lighting condition (with no exposure to the other) performed poorly when tested in the alternate condition (see Table~\ref{tab:results-pure}). We hypothesize that this is due to the model overfitting to the specific characteristics of the training data, which are not present in the opposite lighting condition.

\begin{table}[htbp]
  \centering
  \small
  \setlength{\tabcolsep}{5pt}
  \caption{Steering regression performance by lighting condition, sensor, and training subset (lower RMSE, higher EVA are better).}
  \label{tab:results-pure}
  \begin{tabular*}{\columnwidth}{@{\extracolsep{\fill}}lllrr@{}}
    \toprule
    Lighting               & Sensor               & Train subset & RMSE    & EVA    \\
    \midrule
    \multirow{4}{*}{Day}   & \multirow{2}{*}{DVS} & Day        & 11.75   & 0.690  \\
    &                      & Night      & 22.05   & -0.093 \\
    \cmidrule{2-5}
    & \multirow{2}{*}{APS} & Day        & 16.31   & 0.402  \\
    &                      & Night      & 21.36   & -0.026 \\
    \midrule
    \multirow{4}{*}{Night} & \multirow{2}{*}{DVS} & Day        & 20.78   & 0.026  \\
    &                      & Night      & 7.13    & 0.885  \\
    \cmidrule{2-5}
    & \multirow{2}{*}{APS} & Day        & 21.36   & -0.030 \\
    &                      & Night      & 13.12   & 0.612  \\
    \bottomrule
  \end{tabular*}
\end{table}

\begin{figure*}[htbp]
  \centering
  \def\imgh{0.20\textheight}

  \subfloat[DVS -- Day (rec.\ 1501288723)]{%
    \includegraphics[height=\imgh,keepaspectratio]{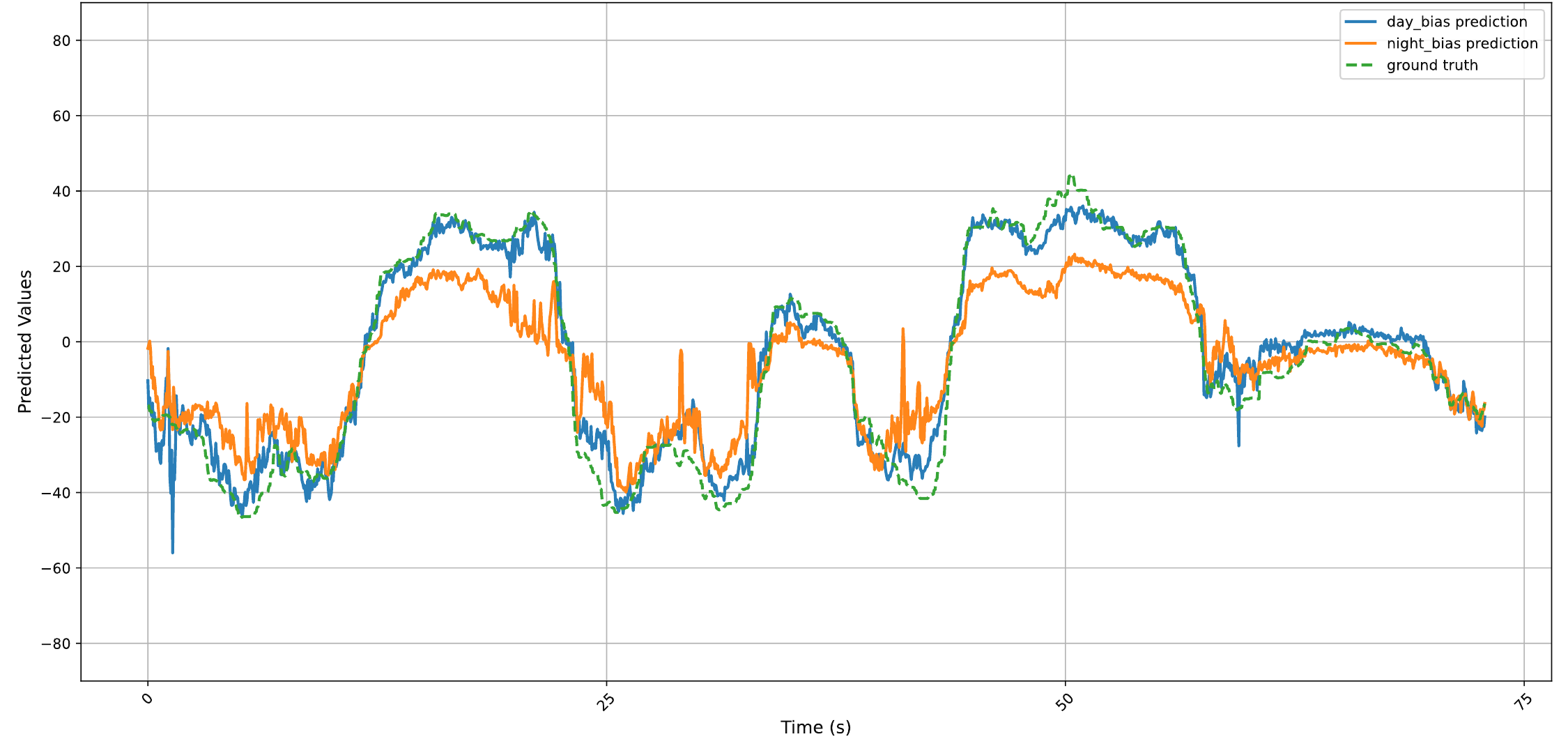}%
  }\hfill
  \subfloat[APS -- Day (rec.\ 1501288723)]{%
    \includegraphics[height=\imgh,keepaspectratio]{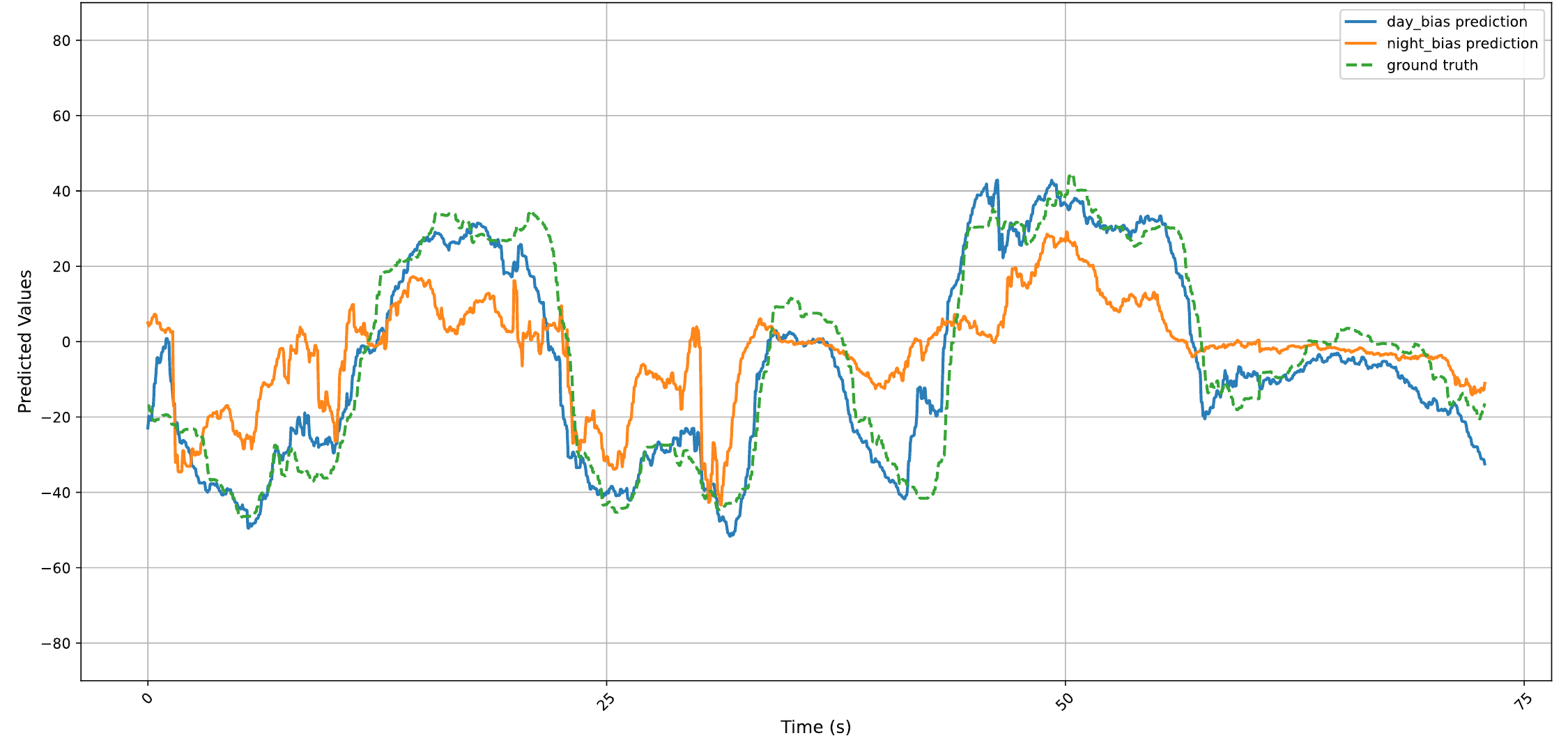}%
  }\hspace{8pt}%
  \subfloat[DVS -- Night (rec.\ 1499656391)]{%
    \includegraphics[height=\imgh,keepaspectratio]{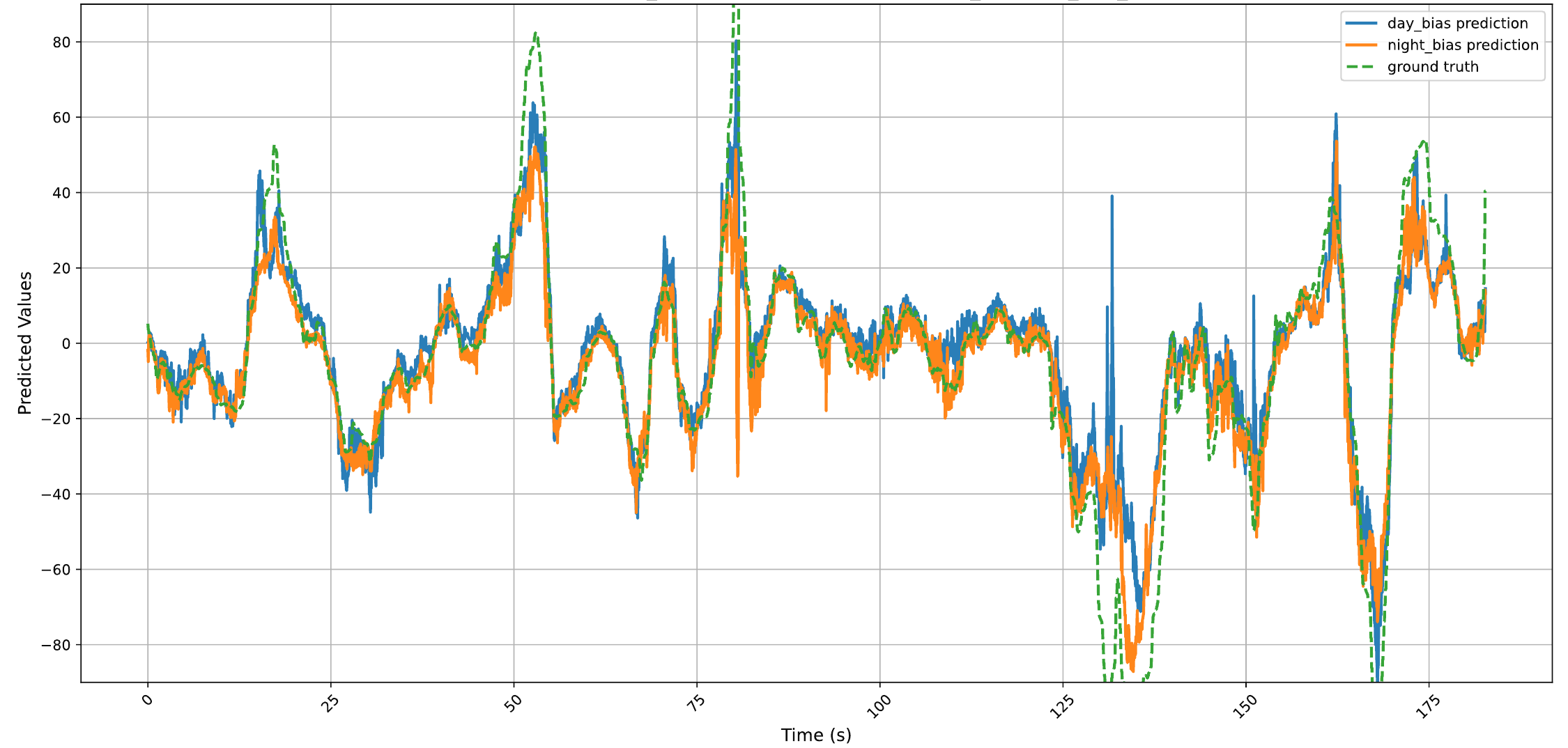}%
  }\hfill
  \subfloat[APS -- Night (rec.\ 1499656391)]{%
    \includegraphics[height=\imgh,keepaspectratio]{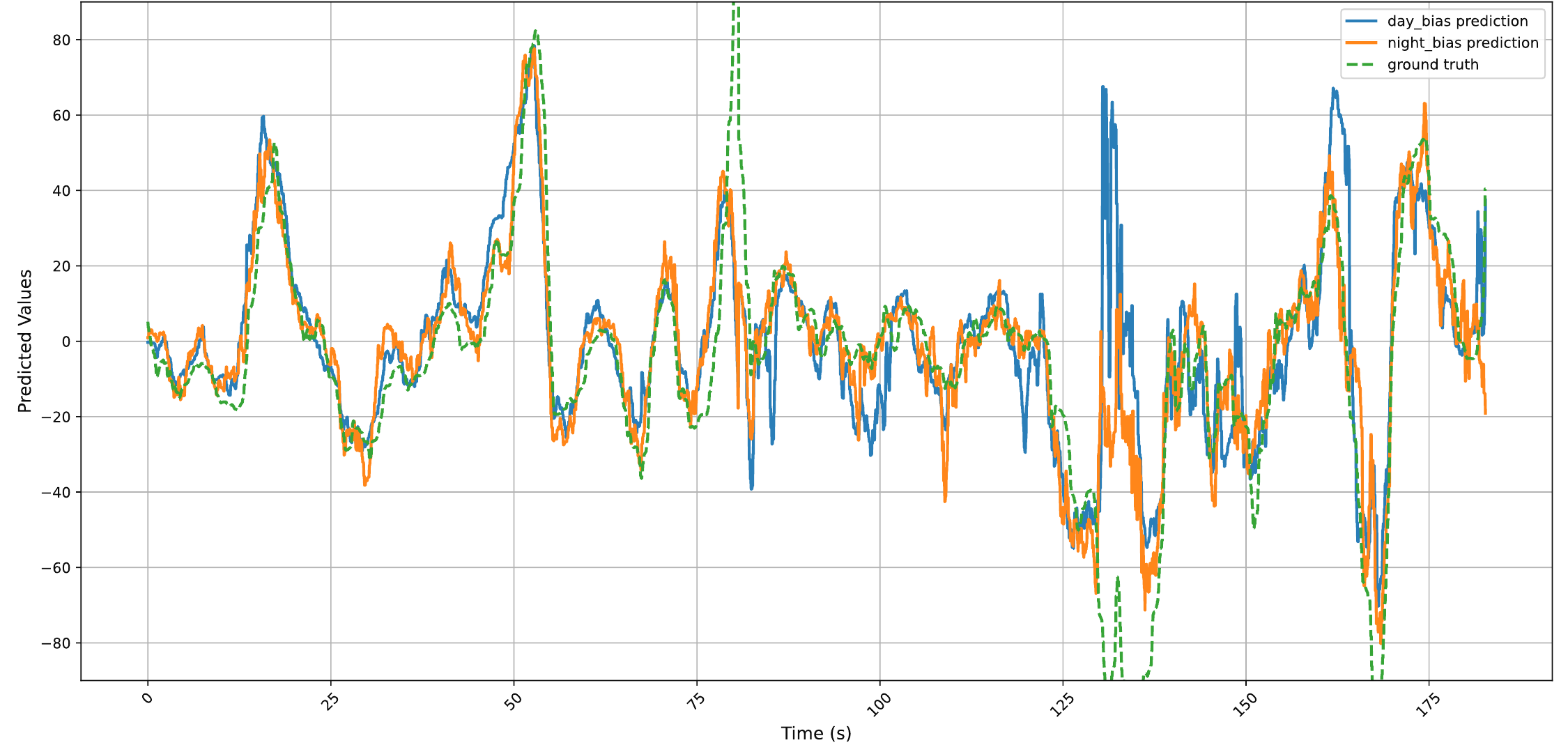}%
  }

  \caption{Comparison across sensors and lighting conditions. The y-axis represents values in degrees of steering. Subfigures (a) and (b) as well as (c) and (d) are from the same recording. The overall better DVS performance, as well as the superior cross-domain performance of the DVS over the APS is visible.}
  \label{fig:dvs_aps_single_column}
\end{figure*}

Following from the results in Table~\ref{tab:results-raw}, we present the domain-shift penalty experienced per-sensor in Table~\ref{tab:results-gap}. While in percentage RMSE changes remain similar across sensor modalities, the DVS maintains higher absolute performances and experiences a smaller percentage drop in EVA, especially when benchmarking the day-biased model in night conditions.
An example comparing the performance of the DVS and APS across lighting conditions can be seen in Figure~\ref{fig:dvs_aps_single_column}.

\begin{table}[htbp]
  \centering
  \small
  \setlength{\tabcolsep}{6pt}
  \caption{Domain-shift penalty within each test condition (mismatch - match, lower RMSE, higher EVA are better).}
  \label{tab:results-gap}
  \begin{tabular*}{\columnwidth}{@{\extracolsep{\fill}}llrr@{}}
    \toprule
    Lighting               & Sensor & RMSE                  & EVA                   \\
    \midrule
    \multirow{2}{*}{Day}   & DVS    & $+5.71$ ($+49.2\%$)  & $-0.37$ ($-53.1\%$)  \\
    & APS    & $+2.69$ ($+16.3\%$)  & $-0.22$ ($-55.6\%$)  \\
    \midrule
    \multirow{2}{*}{Night} & DVS    & $+3.70$ ($+45.7\%$)  & $-0.17$ ($-19.5\%$)  \\
    & APS    & $+5.52$ ($+44.0\%$)  & $-0.38$ ($-59.2\%$)  \\
    \bottomrule
  \end{tabular*}
\end{table}

The results presented in this section show that event cameras can maintain more consistent performance across lighting conditions, exhibiting domain-shift penalties that are in absolute terms smaller than grayscale frames and provide superior baseline performance in cross-domain scenarios. This is a promising result, as it shows that event cameras can be used as an additional sensor modality to mitigate the domain gap problem in end-to-end driving, without requiring additional adjustments such as image-to-image translation. This complements the findings in~\cite{huDDD20EndtoEndEvent2020}, which showed the fusing of APS and DVS data can improve performance over using APS data alone, and emphasizes event camera usage in scenarios where lighting conditions vary significantly.


\section{Conclusion and Future Work} \label{conclusion-section}
In this work we have conducted experiments comparing the performance of event cameras with grayscale frame-based cameras by training end-to-end driving models on datasets biased toward daytime and nighttime conditions, evaluating their robustness to illumination-induced domain shifts. In line with prior intuition, DVS-based models exhibited more consistent performance across domains than APS-based models.

While our experiments do not show that DVS alone closes the illumination domain gap or that its absolute errors are acceptable for deployment, they demonstrate the promise event cameras present for computer vision applications susceptible to a lighting-based domain gap, over an end-to-end driving task. They have also reinforced the idea of adding an event camera to a sensory array if it needs to function across different lighting conditions. These findings have important implications for autonomous robotic systems that must maintain consistent performance across diverse operational environments.

In future work, we aim to determine the level of bias that can be supported in a dataset before the model collapses, in addition to creating a transformation suite to amplify the lighting-invariant nature of event camera data.

\section*{Acknowledgments}
This research was supported by the Australian Government Research Training Program Scholarship.
\clearpage
\bibliographystyle{named}
\bibliography{references}


\end{document}